\documentclass{article}

\usepackage{microtype}
\usepackage{graphicx}
\graphicspath{ {./images/} }
\usepackage{subfigure}
\usepackage{booktabs} 

\newcommand*\rot{\rotatebox{90}}

\usepackage{hyperref}

\usepackage[accepted]{icml2021}

\usepackage{changepage}
\usepackage{nicefrac}
\usepackage[T1]{fontenc}
\newcommand{\opt}{\textsc{OPT}}
\newcommand{\Oh}{\mathcal{O}}

\begin{document}

\twocolumn[
\icmltitle{Robust Learning-Augmented Caching: An Experimental Study}

\icmlsetsymbol{equal}{*}

\begin{icmlauthorlist}
\icmlauthor{Jakub Chłędowski}{equal,to}
\icmlauthor{Adam Polak}{equal,goo}
\icmlauthor{Bartosz Szabucki}{equal,to}
\icmlauthor{Konrad Żołna}{equal,to,ed}
\end{icmlauthorlist}

\icmlaffiliation{to}{Jagiellonian University, Kraków, Poland}
\icmlaffiliation{goo}{EPFL, Lausanne, Switzerland}
\icmlaffiliation{ed}{DeepMind, London, United Kingdom}

\icmlcorrespondingauthor{Jakub Chłędowski}{jakub.chledowski@gmail.com}
\icmlcorrespondingauthor{Adam Polak}{adam.polak@epfl.ch}
\icmlcorrespondingauthor{Bartosz Szabucki}{bartosz.szabucki@gmail.com}
\icmlcorrespondingauthor{Konrad Żołna}{konrad.zolna@gmail.com}

\icmlkeywords{Machine Learning, ICML}

\vskip 0.3in
]



\printAffiliationsAndNotice{\icmlEqualContribution} 

\begin{abstract}
Effective caching is crucial for the performance of modern-day computing systems.
A key optimization problem arising in caching -- which item to evict to make room for a~new item -- cannot be optimally solved without knowing the future.
There are many classical approximation algorithms for this problem, but more recently researchers started to successfully apply machine learning to decide what to evict by discovering implicit input patterns and predicting the future.
While machine learning typically does not provide any worst-case guarantees, the new field of learning-augmented algorithms proposes solutions that leverage classical online caching algorithms to make the machine-learned predictors robust.
We are the first to comprehensively evaluate these learning-augmented algorithms on real-world caching datasets and state-of-the-art machine-learned predictors.
We show that a~straightforward method -- blindly following either a~predictor or a~classical robust algorithm, and switching whenever one becomes worse than the other -- has only a~low overhead over a~well-performing predictor, while competing with classical methods when the coupled predictor fails, thus providing a~cheap worst-case insurance.
\end{abstract}

\section{Introduction}

Caching is an important part of almost any modern-day computing system because it can vastly speed up memory access.
A major optimization problem arising with regard to caching is: \emph{Which item to evict from the cache in order to make room for a~new item when the cache is full?}
The optimization goal is to maximize the number of \emph{cache hits}, i.e.,~situations when the requested item is still present in the cache. If we choose a~wrong item to evict and it is requested again soon after, a~\emph{cache miss} occurs, and the item has to be reloaded to the cache from the main memory\footnote{Even if a requested item is unlikely to be ever reused, it has to be put into the cache.}, which is usually orders of magnitude slower than reading it directly from the cache.

\paragraph{Classical caching algorithms.}
In the \emph{offline} scenario, i.e.,~when we know in advance the sequence of requested items, the problem is easy to solve optimally. Indeed, \citet{Belady66} proved that the number of cache misses is minimized by a~greedy eviction policy -- always evict the item which will reappear the furthest in the future (or which will never reappear, if there is such item).

In the more realistic \emph{online} scenario, we do not know the future requests. 
For the cache size of $k$ items, the classical \textsc{marker} algorithm~\citep{FiatKLMSY91} is $\Oh(\log k)$-competitive, i.e.~it incurs at most $\Oh(\log k \cdot \opt)$ cache misses on inputs which the optimal offline algorithm serves with $\opt$ misses. There is also a~matching lower bound~\citep{FiatKLMSY91}, showing that no algorithm can do better, up to a~constant factor.
On the other hand, real-world applications employ simple heuristics, such as the gold standard Least Recently Used (\textsc{lru}), which happens to perform better in practice.

Recent machine learning approaches try to discover implicit access patterns specific to individual applications, which are likely to occur in future request sequences and use that knowledge to make better eviction decisions~\citep{JainL16,ShiHJL19,LiuHSRA20,YanL20}.

\paragraph{Learning-augmented caching algorithms.}
In general, algorithms based on machine learning models tend to work well on typical inputs but can perform arbitrarily badly when, e.g. training data is scarce, or input patterns change unexpectedly over time. \citet{LykourisV18} proposed a~workaround to that issue. Their online caching algorithm, PredictiveMarker, takes as an additional input for each requested item a~prediction (e.g.~generated by an ML model) when this item will be requested again. The algorithm is \emph{consistent}, i.e.,~it incurs an almost optimal number of cache misses when given nearly perfect predictions, and \emph{robust}, i.e.,~it is $O(\log k)$-competitive (just like the optimal \textsc{marker} algorithm) even when the predictions are completely wrong.\footnote{The exact meaning of consistency and robustness seems to be somewhat inconsistent throughout the literature.} Formally, PredictiveMarker incurs at most
\[\Oh\left(\opt \cdot \min\left(\sqrt{\frac{\eta_{\mathrm{reuse}}}{\opt}} + 2, \log k\right)\right)\]
cache misses, where $\eta_{\mathrm{reuse}}$ denotes the total $L1$ error of the predictor.
\citet{Rohatgi20} and \citet{Wei20} came up with more caching algorithms working in this setup, with improved dependencies on the prediction error $\eta_{\mathrm{reuse}}$.

\citet{AntoniadisCE0S20} proposed a~different setup for learning-augmented caching algorithms. In their setup, a~predictor has to keep guessing what the optimal (knowing the future) policy would do. In principle, any caching algorithm can serve as a~predictor itself. The prediction error $\eta_{\mathrm{cache}}$ is measured as the size of the symmetric difference between cache configurations of the optimal algorithm and the predictor, summed over all time steps. Actually, the setup works for any \emph{metrical task system} (MTS), a~general class of online problems that includes caching. \citet{AntoniadisCE0S20} provide two consistent and robust algorithms (one deterministic and one randomized) for general MTS, and specifically for caching, they propose an algorithm, dubbed Trust\&Doubt, with a~better dependence on prediction error.

\paragraph{Motivation.}

A comprehensive comparison of the above learning-augmented caching algorithms is a~difficult task. Admittedly, worst-case competitive ratios of algorithms within the same setup can be compared, but such theoretical comparison between the two setups is implausible. It follows from the fact that the final competitive ratios depend on coupled predictors, with different error measures that can not be translated between the setups.

To the best of our knowledge, so far there is no experimental evaluation of these algorithms using real-world datasets nor predictors. Experiments were included only in works by \citet{LykourisV18} and \citet{AntoniadisCE0S20}. However, these are small-size proof-of-concept experiments on datasets adapted from other problems (not related to caching) and using simple ad-hoc predictors instead of fully-fledged machine learning models. The following question remains wide open.

\begin{center}
\emph{Are learning-augmented caching algorithms practical?}
\end{center}

\paragraph{Our study.}
In this paper, we set out to answer this question experimentally. We use benchmark dataset from the 2nd Cache Replacement Championship~\citep{CRC17}, also used by \citet{ShiHJL19} and \citet{LiuHSRA20}. As a~predictor, we use state-of-the-art machine-learning-based caching algorithm Parrot~\citep{LiuHSRA20}.
Conveniently, Parrot also predicts when items reappear in the request sequence, on top of predicting which item the optimal policy discards.
It can thus be used as a~predictor for learning-augmented algorithms in both setups, by \citet{LykourisV18} and by \citet{AntoniadisCE0S20}, respectively.

First, we test how the algorithms perform with a~fully-fledged predictor so that we can see if the overhead they incur is an acceptable cost to pay in exchange for the worst-case guarantees they provide. Second, we also run a~scenario with a~predictor under-performing due to the~scarcity of available training data. That lets us evaluate if the theoretical robustness guarantees play a~role in practice.

Our secondary contribution is a~ready-to-use benchmark, which facilitates an easier future comparison with our results. It is based on a~dataset which was previously used by \citet{LiuHSRA20}. However, since the method they use to generate input instances tends to have high variance, by providing the ready-to-use inputs we make it possible to compare directly with the numbers we report, without having to rerun all the experiments.

\section{Background and Algorithms}

In this section, we aim to concisely describe the background needed for the understanding of the following sections.
We will start by explaining the two prediction setups for learning-augmented caching.
Next, we will continue with a short overview of the learning-augmented algorithms used in each setup.
At the end, we will describe Parrot~\citep{LiuHSRA20}, the neural network predictor that we use to generate predictions for the algorithms.

\begin{table*}
\caption{{\bf Classical and learning-augmented caching algorithms.} Constants in competitive ratios are omitted for brevity.}
\label{algorithms-table}
\begin{center}
\begin{small}
\begin{tabular}{@{}lllll@{}}
\toprule
Algorithm & Prediction & Competitive ratio & Combiner & Reference \\
\midrule
\textsc{opt}           & n/a            & $1$    & n/a & \citet{Belady66} \\
\textsc{lru}                & n/a            & $k$   & n/a & folklore \\
\textsc{marker}             & n/a            & $\log k$ & n/a & \citet{FiatKLMSY91} \\
\addlinespace \midrule
PredictiveMarker  & reuse distance & $\min(\log k, \sqrt{\eta_{\mathrm{reuse}}/\opt}) $ & n/a & \citet{LykourisV18} \\
LMarker           & reuse distance & $\min(\log k, \log(\eta_{\mathrm{reuse}}/\opt)) $ & n/a & \citet{Rohatgi20} \\
LNonMarker$^\mathrm{D}$         & reuse distance & $\min(\log k, \nicefrac{\log k}{k} \cdot \eta_{\mathrm{reuse}}/\opt)$ & deterministic & \citet{Rohatgi20} \\
LNonMarker$^\mathrm{R}$         & reuse distance & $\min(\log k, \nicefrac{\log k}{k} \cdot \eta_{\mathrm{reuse}}/\opt)$ & randomized & \citet{Rohatgi20} \\
BlindOracle$^\mathrm{D}$      & reuse distance & $\min(\log k, \nicefrac{1}{k} \cdot
\eta_{\mathrm{reuse}}/\opt)$ & deterministic & \citet{Wei20} \\
BlindOracle$^\mathrm{R}$      & reuse distance & $\min(\log k, \nicefrac{1}{k} \cdot
\eta_{\mathrm{reuse}}/\opt)$ & randomized & \citet{Wei20} \\
\addlinespace \midrule
RobustFtP$^\mathrm{D}$                & optimal policy & $\min(\log k, \eta_{\mathrm{cache}} / \opt)$ & deterministic & \citet{AntoniadisCE0S20} \\
RobustFtP$^\mathrm{R}$                & optimal policy & $\min(\log k, \eta_{\mathrm{cache}} / \opt)$ & randomized & \citet{AntoniadisCE0S20} \\
Trust\&Doubt            & optimal policy & $\min(\log k, \log(\eta_{\mathrm{cache}}/\opt)) $ & n/a & \citet{AntoniadisCE0S20} \\
\bottomrule
\end{tabular}
\end{small}
\end{center}
\vskip -0.1in
\end{table*}

\subsection{Prediction setups for caching}

\citet{LykourisV18} proposed the first prediction setup for online caching. In their setup, after each request, the predictor has to forecast when the requested item will be requested again -- the value called \emph{reuse distance}. This is a~very natural choice since the reuse distance is the only statistic that Belady's optimal offline algorithm looks at. They define the prediction error $\eta_{\mathrm{reuse}}$ to be the $L1$ distance, i.e.,~the sum over all requests of the absolute difference between the actual and predicted reuse distances.

\citet{AntoniadisCE0S20} noticed a~limitation of this setup -- it does not generalize to other online problems. Already for the weighted caching problem (where the cost of loading each item can be different) even perfect reuse distance predictions are not sufficient to beat the best classical prediction-less algorithm.
To address this limitation, they proposed an alternative setup that works for \emph{metrical task systems} -- a~general class of online problems, which includes caching. In their setup, after each request, the predictor has to guess what an optimal offline algorithm would do. The prediction error $\eta_{\mathrm{cache}}$ is the size of the symmetric difference between the caches maintained by the optimal algorithm and the predictor, summed over time.

A direct comparison of the two setups is problematic for at least two reasons. First, even though predictions for the first setup can be translated (by following Belady's rule) to predictions for the second setup (but not the other way round), the respective errors cannot be related to each other, as shown by the two instructive examples in \citet[Sect.~1.3]{AntoniadisCE0S20}. Second, a~priori, it is not clear which of the two types of predictors is easier to train well. On the one hand, predicting reuse distances can be framed as a~standard supervised learning task, while predicting optimal policy seems to require more advanced approaches such as imitation learning or reinforcement learning. On the other hand, one can imagine an input distribution such that it is hard to accurately predict reuse distances while it is still easy to always find an item with a~reuse distance likely so large that it is safe to evict.

\begin{table*}
\caption{\textbf{Characteristics of our datasets.} The first row shows the total sizes of all used datasets. These sizes are later split into train/valid/test sets with 80\%/10\%/10\% splits. Further rows display the cache hit rates of pure (non-learning-augmented) algorithms, illustrating varying difficulties of the datasets.}
\label{reevaluation-table}
\begin{center}
\begin{small}
\resizebox{\textwidth}{!}{\begin{tabular}{@{}lccccccccccccc@{}}
\toprule
  & \textbf{astar} & \textbf{bwaves} & \textbf{bzip} & \textbf{cactusadm} & \textbf{gems} & \textbf{lbm} & \textbf{leslie3d} & \textbf{libq} & \textbf{mcf} & \textbf{milc} & \textbf{omnetpp} & \textbf{sphinx3} & \textbf{xalanc} \\
\midrule
\textbf{Size} & 1,154,048 & 570,368 & 167,680 & 221,952 & 723,456 & 782,080 & 716,032 & 579,840 & 2,965,504 & 556,800 & 555,520 & 328,704 & 69,120 \\
\textbf{Cache hits} \\
\quad \textsc{opt} & 37.4\% & 4.9\% & 80.8\% & 33.7\% & 12.7\% & 24.8\% & 30.9\% & 5.3\% & 44.6\% & 1.4\% & 42.4\% & 74.8\% & 56.9\%  \\
\quad \textsc{random} & 8.3\% & 0.2\% & 56.5\% & 4.5\% & 3.9\% & 2.2\% & 9.5\% & 0.0\% & 20.5\% & 0.0\% & 17.6\% & 52.8\% & 36.8\% \\
\quad \textsc{lru} & 4.0\% & 0.0\% & 63.8\% & 0.0\% & 2.9\% & 0.0\% & 9.5\% & 0.0\% & 27.1\% & 0.0\% & 20.4\% & 12.7\% & 45.4\%  \\
\quad \textsc{marker} & 4.7\% & 0.0\% & 62.7\% & 1.3\% & 4.1\% & 0.0\% & 9.3\% & 0.0\% & 24.9\% & 0.0\% & 20.3\% & 42.2\% & 43.5\%  \\
\quad \textsc{parrot-reuse} & 29.0\% & 0.1\% & 57.9\% & 21.8\% & 0.4\% & 0.5\% & 4.9\% & 5.3\% & 32.5\% & 1.1\% & 11.9\% & 67.6\% & 37.3\%  \\
\quad \textsc{parrot-cache} & 32.2\% & 0.3\% & 68.4\% & 32.9\% & 3.1\% & 0.0\% & 11.4\% & 0.0\% & 43.9\% & 0.0\% & 21.9\% & 70.6\% & 49.7\%  \\
\bottomrule
\end{tabular}}
\end{small}
\end{center}
\vskip -0.1in
\end{table*}

\subsection{Augmented algorithms}
\label{sec:augmented}

We evaluate the six learning-augmented caching algorithms proposed up to date. PredictiveMarker~\citep{LykourisV18}, LMarker and LNonMarker~\citep{Rohatgi20}, and BlindOracle~\citep{Wei20} are all augmented with reuse distance predictions, while RobustFtP and Trust\&Doubt~\citep{AntoniadisCE0S20} utilize optimal policy predictions. See Table~\ref{algorithms-table} for a~summary of these algorithms.

\paragraph{Combiners.}
One way to achieve robustness is to combine a~non-robust learning-augmented algorithm with a~robust classical algorithm, e.g.~\textsc{marker}, using a~\emph{combiner}. A~combiner is a~procedure that takes two algorithms and uses them in a~black-box way to perform on par with the better of the two algorithms on each input (up to a~constant~factor).

A straightforward combiner can simulate the two algorithms, keep track of their respective costs up to date, and switch between them whenever one heavily outperforms the other. The idea dates back to \citet{FiatRR94}, and it was adapted to the learning-augmented setting by \citet{LykourisV18}. We will call the above combiner \emph{deterministic}. In contrast, \citet{Wei20} and \citet{AntoniadisCE0S20} use an idea of \citet{BlumB00} to provide an alternative \emph{randomized} combiner. The more intricate combining algorithm allows to bring down the multiplicative overhead to $1+\varepsilon$ at the cost of an extra additive constant depending on $\varepsilon$.

Some learning-augmented algorithms are built using combiners, while others achieve robustness out-of-the-box. In principle, each of the algorithms implementing a~combiner can have (at least) four variants. It can be paired with either \textsc{marker} or \textsc{lru}\footnote{Formally, combining with \textsc{lru} does not yield robustness, since \textsc{lru} is not $\Oh(\log k)$-competitive. However, it has at least a provably bounded competitive ratio, and its good practical performance is well understood. Hence, in practice, it makes sense to use \textsc{lru} as a~fallback option for a~potentially arbitrarily inaccurate machine-learned algorithm.}, and, independently of that choice, it can use either the deterministic or the randomized combiner.

\paragraph{Algorithms with reuse distance.}
PredictiveMarker, LMarker, and LNonMarker all work by keeping track of \emph{eviction chains}. An eviction chain is a~sequence of sub-optimal evictions where each eviction is forced by the cache-miss that can be blamed on the previous eviction in the sequence. The three algorithms differ with respect to (1) for how long they trust predictions in each eviction chain, and (2) what they do instead when they eventually lose the trust. We refer to the original papers for detailed discussions of these strategies, designed to allow better and better for worst-case competitive ratios, expressed as a~function of the normalized error $\eta_{\mathrm{reuse}}/\mathrm{OPT}$. PredictiveMarker and LMarker are provably robust out-of-the-box. In contrast, LNonMarker achieves robustness thanks to the classical \textsc{marker} algorithm and a~black-box combiner. 

The next learning-augmented caching algorithm, BlindOracle, simply applies Belady's rule to the prediction -- i.e.~it evicts the item with the predicted next arrival time furthest in the future -- and adds the guarantee of a~robust algorithm (e.g.~\textsc{marker}) using either of the aforementioned combiners. We note that, despite being the simplest out of the four, BlindOracle also has the best competitive ratio, though the proof is much more involved than the algorithm itself.

\paragraph{Algorithms with policy predictions.}
RobustFtP stands for \emph{Robust Follow-the-Prediction}. It is based on essentially the same algorithmic idea as BlindOracle -- i.e.~apply a combiner to (1) robust \textsc{marker} and (2) consistent following the predictions -- but in the alternative prediction setup and with a~quite different theoretical analysis.

Trust\&Doubt utilizes predictions in a~more intricate way in order to achieve better dependence on the prediction error $\eta_{\mathrm{cache}}$. Similar to PredictiveMarker and LMarker, it is robust out-of-the-box and hence does not require a~combiner.

\subsection{Predictor}
We couple the learning-augmented algorithms with the state-of-the-art predictors. Specifically, \citet{LiuHSRA20} proposed Parrot, an imitation learning algorithm that tries to mimic Belady's oracle policy with a deep neural network.

At each step, the model is input the last $H=30$ requested items (including the current one) and the $k=16$ items that are at the time in the cache (in the relevant set).
All these items are encoded to fixed-size embedding, and the requested items are additionally processed by LSTM~\cite{lstm}, which is continuously fed with them one by one.
That leads to obtaining $H + k$ embeddings, which are further processed with attention (specifically, Transformer~\citep{Vaswani17} and BiDAF~\citep{seo2018bidirectional}), which finally outputs one vector per cache item.
These elements are passed through a~linear layer with softmax to form a~prediction.

The model uses two different prediction heads, which are optimized simultaneously during training.
The first one predicts which element would be evicted from the cache by Belady's oracle and defines the \textsc{parrot-cache} predictor, which is exactly what the RobustFtP and Trust\&Doubt algorithms need.

The second head estimates, for each element in the cache, the number of steps before the element is requested again. 
We use that estimate for the last requested item to construct the second predictor, which we call \textsc{parrot-reuse}.
Such predictions are compatible with the PredictiveMarker, LMarker, LNonMarker, and BlindOracle algorithms.

\section{Datasets}

Our datasets come from the 2nd Cache Replacement Championship~\citep{CRC17} and consist of real-world memory access traces from the SPEC CPU2006 benchmark~\citep{henning2006}.

\enlargethispage{\baselineskip}  

There are two ways to obtain the traces to begin with. One can either download the traces released in the Cache Replacement Championship or collect custom traces using a~dynamic binary instrumentation tool DynamoRIO~\citep{bruening2003}. \citet{LiuHSRA20} used the second method to evaluate Parrot and shared their procedure, but they were unable to release their exact traces. We follow this procedure to create datasets from the publicly available traces~\citep{CRC17}. In particular, we subsample the traces by choosing 64 out of 2048 \emph{sets}\footnote{A \emph{set-associative} cache is split into $n$ sets, each holding $k$ items, called \emph{lines}. Each line's address uniquely predetermines the set it can be cached in. In that sense, each trace constitutes $n$ independent instances of the caching problem. (Here $k=16$.)} and filtering the accesses\footnote{Following \citet{ShiHJL19} and \citet{LiuHSRA20} we evaluate our approach on the last level of a~three-level cache hierarchy.} to those sets. The first 80\% of this sequence is used for training, followed by 10\% used for validation, and the last 10\% for testing. We refer to Table~\ref{reevaluation-table} for details of the datasets.

We noticed that even slight differences in the data collection and postprocessing might create datasets of significantly different characteristics. To account for that, we have reevaluated Parrot on our inputs, and made them publicly available at
\href{https://github.com/chledowski/Robust-Learning-Augmented-Caching-An-Experimental-Study-Datasets}{https://github.com/chledowski/Robust-Learning-Augmented-Caching-An-Experimental-Study-Datasets}.

\section{Experimental Setup}

For both Parrot and the learning-augmented algorithms, we use the code made available by \citet{LiuHSRA20} and \citet{AntoniadisCE0S20}, respectively.
We unified both codebases and connected them into a single pipeline.
The source code is publicly available at \href{https://github.com/chledowski/ml\_caching\_with\_guarantees}{https://github.com/chledowski/ml\_caching\_with\_guarantees}.

Due to constrained computing resources, we limited the number of training steps to $20\,000$.
The second change is to ablate DAgger~\citep{pmlr-v15-ross11a}, as it did not yield any improvements in our case.
No other Parrot's hyper-parameter was changed. 

As a result, our models are trained for $20\,000$ steps on each dataset, with a batch size of $32$.
The best model is chosen to be the one with the highest validation cache hit rate among the evaluated checkpoints (done every $5000$ steps).

To measure the practicality of learning-augmented caching algorithms, we will weaken the Parrot model by training only on prefixes (e.g.~1\%) of the original datasets.
We will analyze how the learning-augmented algorithms perform under such a change.

As we mention in Section~\ref{sec:augmented}, each of the algorithms implementing a~combiner (LNonMarker, BlindOracle, and RobustFtP) have four variants, using the deterministic or the randomized combiner, with \textsc{marker} or with \textsc{lru}.
We noticed that \textsc{marker} and \textsc{lru} perform on par with each other on most of the datasets that we use. 
Hence, for simplicity, we analyze only the two variants with \textsc{marker} (deterministic and randomized).
For completeness, we also ran experiments with \textsc{lru}, but as expected the choice turns out to be irrelevant.

\begin{figure*}[t!]
    \centering
    \includegraphics[width=1\textwidth]{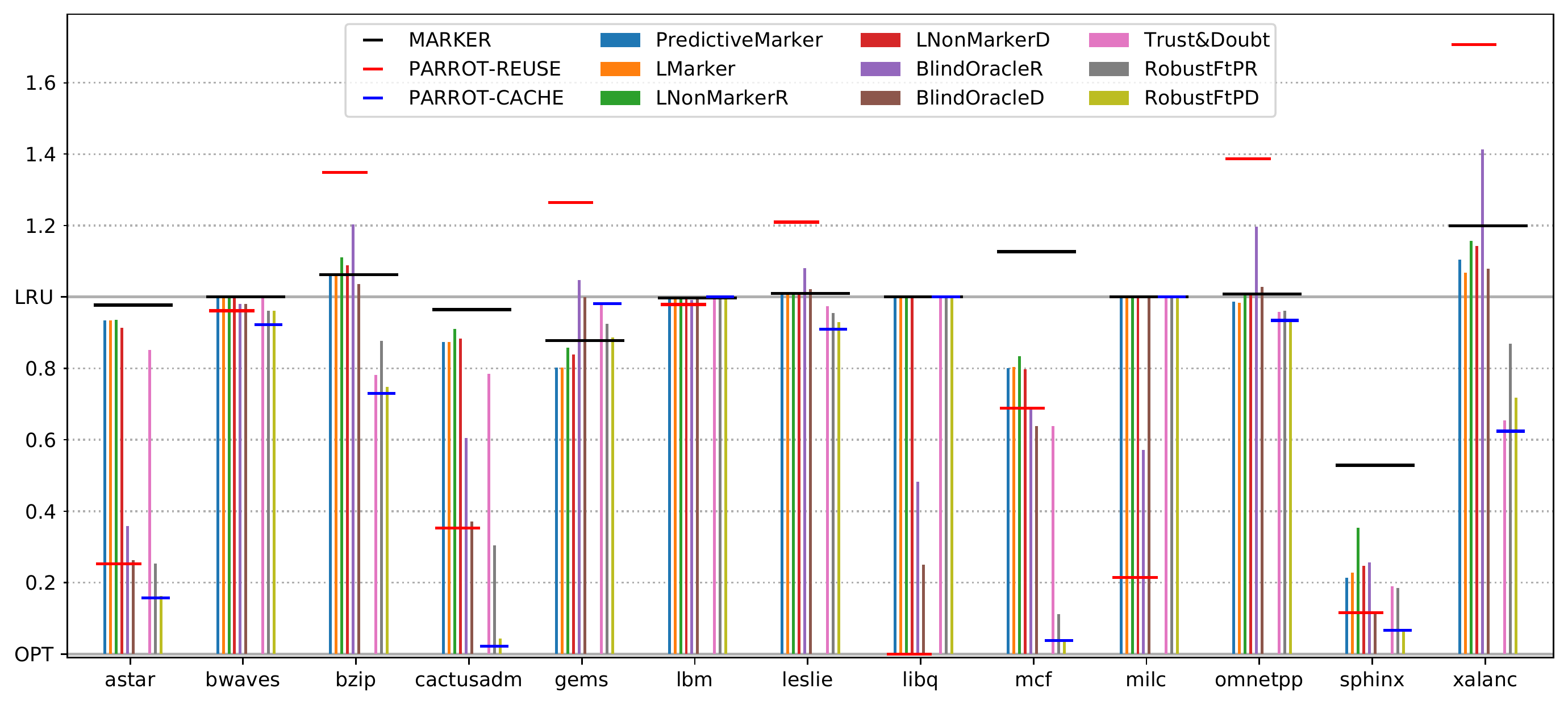}
    \vspace{-0.4cm}
    \caption{{\bf Normalized cost (the lower the better) of learning-augmented algorithms compared to coupled predictors and classical algorithms.} The bars reflect the performance of the augmented algorithms, while horizontal lines correspond to non-augmented methods. The scores reflect LRU-normalized empirical competitive ratio (defined in Section~\ref{sec:normalize}). The predictor \textsc{parrot-cache} (blue line) outperforms \textsc{parrot-reuse} (red line) and even approaches optimal policy for a few datasets. \textsc{marker} (black line) is comparable to \textsc{lru}. The best augmented methods are RobustFtP$^\mathrm{D}$ (yellow bar) and BlindOracle$^\mathrm{D}$ (brown bar) which have very low overhead while stay robust when the coupled predictor performs worse than \textsc{marker}. RobustFtP$^\mathrm{D}$ is, however, coupled with better predictors and hence significantly outperforms BlindOracle$^\mathrm{D}$. The remaining augmented algorithms are overly conservative and hence fail to leverage good predictions.
    \label{fig:results_full}}
    \vspace{0.8cm}
    \centering
    \includegraphics[width=\textwidth]{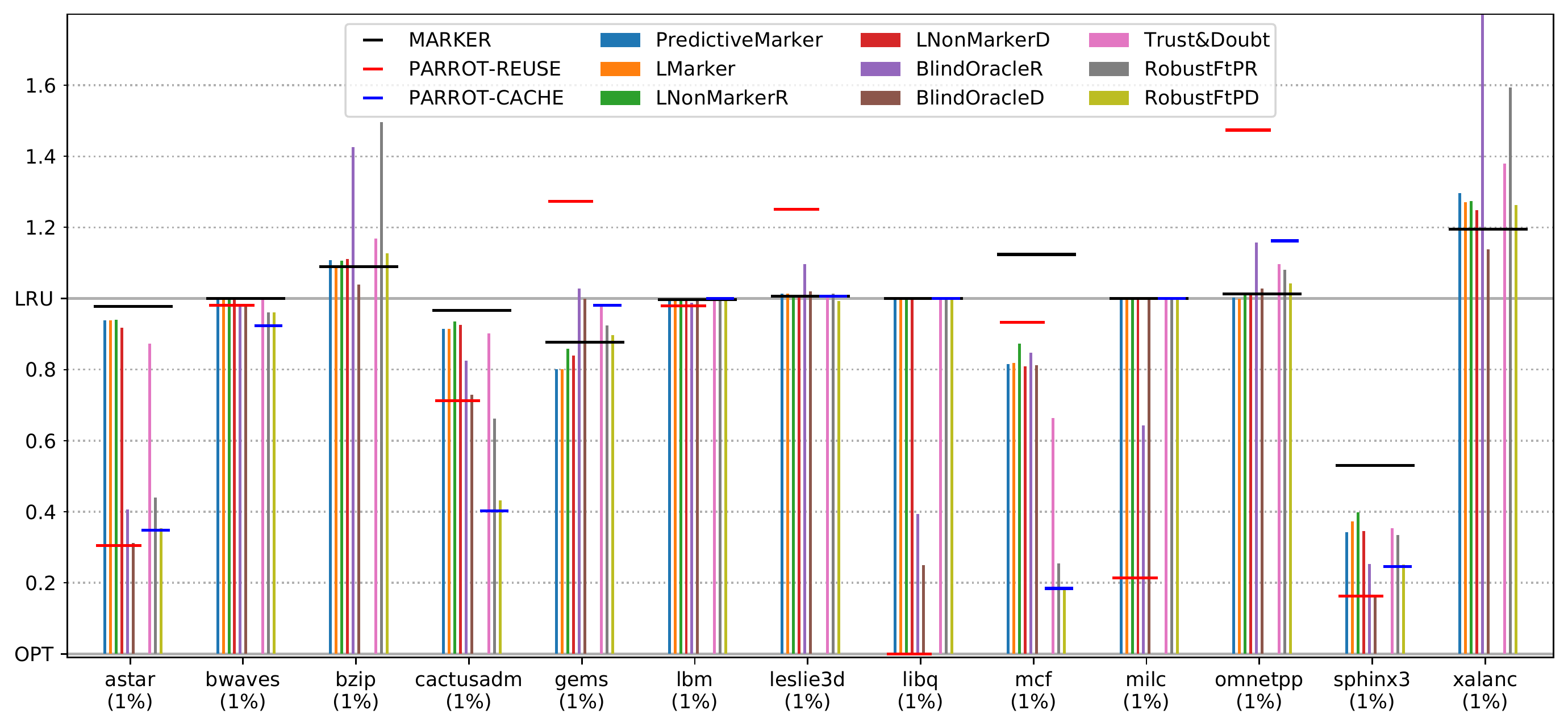}
    \vspace{-0.4cm}
    \caption{{\bf Normalized cost (the lower the better). Underperforming predictors trained on 1\% of data.} The predictors (red and blue lines) significantly outperform \textsc{marker} (black line) for only 4 datasets due to limited training data. Augmented algorithms (bars) are robust -- they approach \textsc{marker} (black line), even when coupled predictors are wildly inaccurate. The robustness of RobustFtP$^\mathrm{R}$ (violet bar) and BlindOracle$^\mathrm{R}$ (gray bar) seems the weakest (especially for bzip and xalanc datasets). At the same time, RobustFtP$^\mathrm{D}$ (yellow bar) and BlindOracle$^\mathrm{D}$ (brown bar) prove to be the best, as they are able to leverage accurate predictions, while the remaining augmented algorithms are overly conservative. Predictors' scores for bzip and xalanc are not shown as they are above 2.0. Most augmented algorithms, however, still perform comparably to \textsc{marker} for these two datasets.
    \label{fig:results_1p}}
\end{figure*}

\begin{figure*}[t]
    \centering
    \includegraphics[trim={0 12cm 0 0},clip,width=\textwidth]{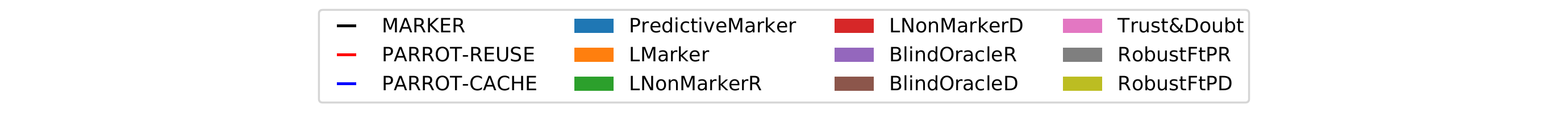}
    \vspace{0.1cm}
    \begin{minipage}[t]{0.486\textwidth}
    \includegraphics[width=1\columnwidth]{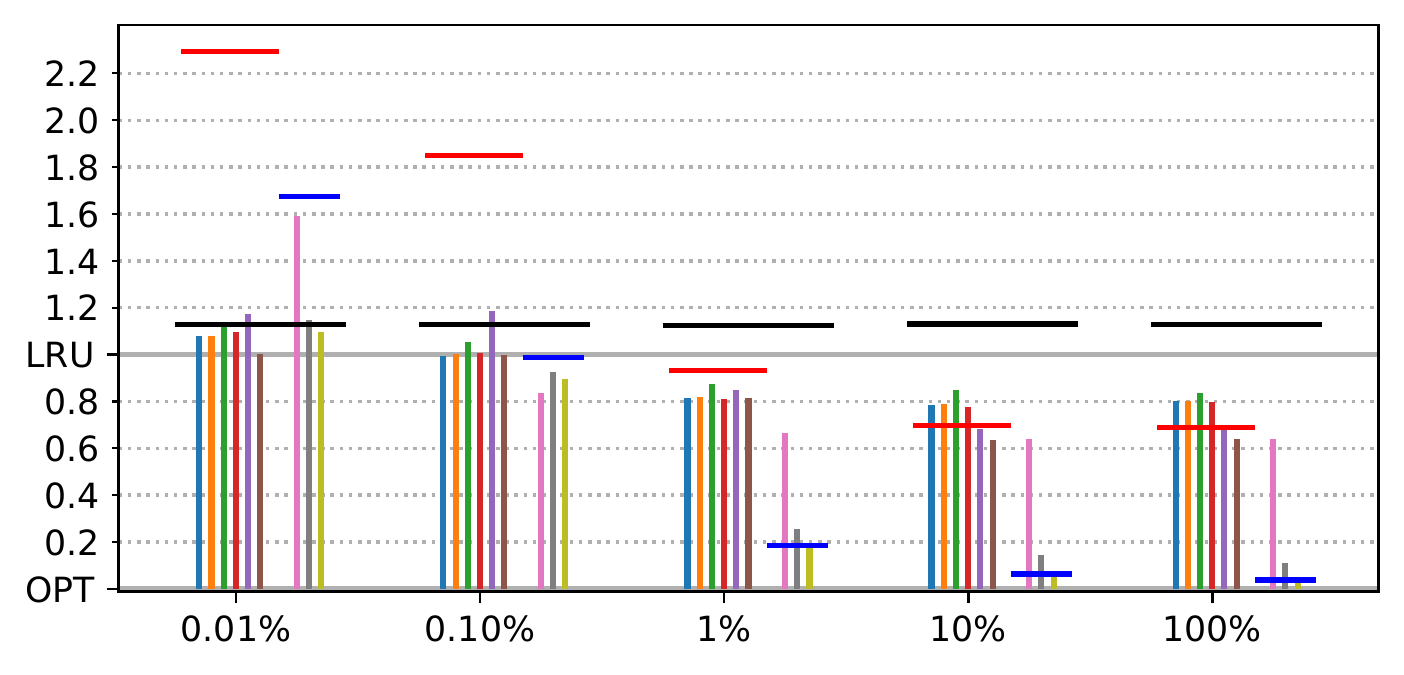}
    \vspace{-0.4cm}
    \caption{{\bf Normalized cost (the lower the better). A closer look at mcf subsamples.} Performance of the \textsc{parrot-reuse} and \textsc{parrot-cache} predictors improve along with the growing size of the trained dataset, while the two best augmented algorithms in the respective setups are able to follow the minimum of the results of the combined predictor and classical algorithms.
    \label{fig:results_mcf}}
    \end{minipage}
    \hfill
    \begin{minipage}[t]{0.486\textwidth}
    \includegraphics[width=1\columnwidth]{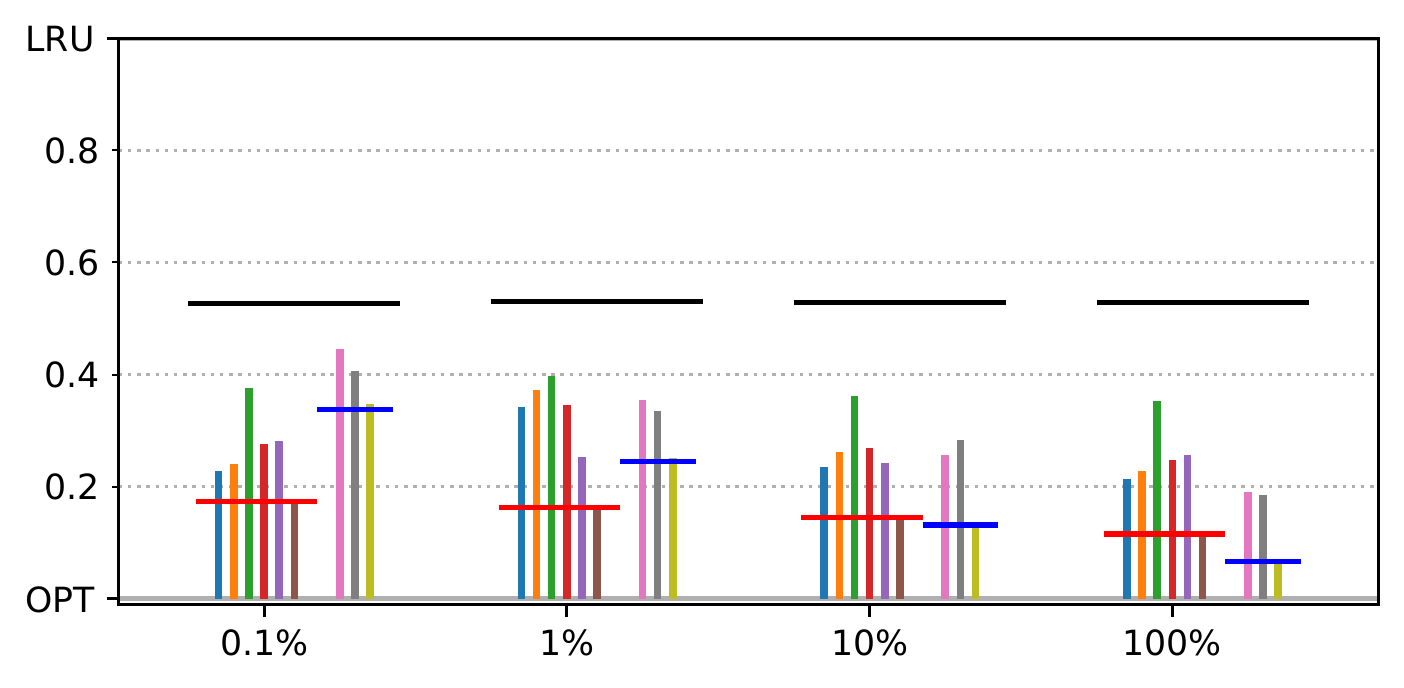}
    \vspace{-0.4cm}
    \caption{{\bf Normalized cost (the lower the better). A closer look at sphinx3 subsamples.} The \textsc{parrot-reuse} is more robust to the scarcity of data when trained on subsamples of sphinx3. The predictors improve along with the growing size of the trained dataset. The augmented algorithms perform significantly better than \textsc{marker}, but only RobustFtP$^\mathrm{D}$ and BlindOracle$^\mathrm{D}$ approach the predictor.
    \label{fig:results_sphinx}}
    \end{minipage}
\end{figure*}

\section{Results}

In this section, we analyze our experiments on the practicality of the learning-augmented algorithms. We begin with full training sets to assess algorithms' overhead over fully-fledged predictors. Then we move to predictors trained on only $1\%$ of data to check whether the algorithms are robust in practice. We end with a closer look at two datasets, which we further subsample to better illustrate the behavior of the algorithms coupled with predictors of varying accuracy.

To make comparison across datasets easier, we normalize scores to both \textsc{lru} and Belady's \textsc{opt} at the same time. Specifically, for an algorithm $ALG$ with empirical competitive ratio\footnote{The \emph{empirical competitive ratio} of the algorithm $ALG$ is defined as $\mathrm{cr}_{ALG} = \mathrm{cost}_{ALG} / \mathrm{cost}_{OPT}$, where $\mathrm{cost}_{ALG}$ denotes the number of cache misses that algorithm $ALG$ incurs on a dataset.} $\mathrm{cr}_{ALG}$, we report the \emph{LRU-normalized empirical competitive ratio}, i.e.,
\[\frac{\mathrm{cr}_{ALG} - 1}{\mathrm{cr}_{LRU} - 1}.\]
The lower the value, the better, meaning that the algorithm's performance is closer to Belady's optimal oracle. For the sake of completeness, the raw unnormalized scores are included in the supplementary material.

\label{sec:normalize}

\subsection{State-of-the-art predictor}

\label{sec:full}

In order to be applicable in practice, a~learning-augmented caching algorithm should have a~low overhead.
In other words, when the underlying predictor is accurate, the final performance should be as close to the predictor's performance as possible.
In our first experiment, we aim to test which of the learning-augmented caching algorithms fulfill this key requirement.

To this end, for each dataset considered, we train the Parrot model on the full training set, which leads to obtaining two state-of-the-art predictors -- \textsc{parrot-reuse} and \textsc{parrot-cache}.
Then, we couple each learning-augmented caching algorithm with one of the predictors, depending on the nature of the algorithm.
The results are presented in Figure~\ref{fig:results_full}.

The bars correspond to the augmented algorithms, while horizontal lines reflect the performance of non-augmented methods.
\textsc{parrot-cache} is almost always better than the classical \textsc{lru} baseline and even approaches optimal behavior for a~few datasets.
\textsc{parrot-reuse} performs significantly worse than \textsc{parrot-cache} and even lags behind \textsc{lru} in a~few cases.
It makes the use of algorithms coupled with \textsc{parrot-cache} preferable, at least with the currently best available predictors.

When a~predictor is better than \textsc{marker}, all learning-augmented algorithms leverage its accurate predictions to improve over the classical baseline.

However, only the two simplest algorithms -- BlindOracle$^\mathrm{D}$ and RobustFtP$^\mathrm{D}$ -- have overheads over predictors low enough to be considered practical.
Note that these two are based on the same rule, just applied in two different settings. The rule is to blindly follow either a~predictor or \textsc{marker}, and switch whenever one heavily outperforms the other (see Section~\ref{sec:augmented}).
The versions of these \mbox{methods} with \mbox{randomized} combinations (i.e.,~BlindOracle$^\mathrm{R}$ and RobustFtP$^\mathrm{R}$) seem to overly rely on \textsc{marker}, and their performance is clearly worse.

All remaining methods offer only a~slight improvement over \textsc{marker}, even if the predictor excels.
These complex strategies on how to use predictions, developed to prove worst-case bounds, do not work in practice for the analyzed cases.

In a~nutshell, \textsc{parrot-cache} significantly outperforms all algorithms and RobustFtP$^\mathrm{D}$ uses it with only a~small overhead.
We will check how these methods work when coupled with underperforming predictors in the next subsection.

\subsection{Underperforming predictor}

As mentioned before, we impair the training procedure to obtain underperforming predictors.
Specifically, we heavily subsample the training sets, leaving the first 1\% of requests available to the models.
The results are presented in Figure~\ref{fig:results_1p}.

Interestingly, even when trained on only 1\% of the data, \textsc{parrot-reuse} and \textsc{parrot-cache} are still sometimes able to find policies better than \textsc{marker}.
In these cases, the best learning-augmented caching algorithms perform better than \textsc{marker}, with a small overhead, similar to the results in the previous subsection.

In most cases, however, the predictors overfit to severely limited training data, and the resulting caching strategy is no better than \textsc{marker}.
Both \textsc{parrot-cache} and \textsc{parrot-reuse} lag behind the classical method for a few tasks (bzip, gems, leslie3d, omnetpp and xalanc).
In these cases, the augmented algorithms -- with the exception of BlindOracle$^\mathrm{R}$ and RobustFtP$^\mathrm{R}$ for two datasets -- perform comparably with \textsc{marker}, which empirically proves their robustness.
Notably, the robustness of the two simplest methods (BlindOracle$^\mathrm{D}$ and RobustFtP$^\mathrm{D}$), shown in the previous subsection to be much better at leveraging good predictions, is comparable to the rest of the algorithms.

In short, together with conclusions from the previous subsection, the two methods closely track the better of their two components -- the predictor or \textsc{marker}.
The remaining methods add too much overhead when coupled with well-performing predictors, while it does not result in better robustness.

\subsection{A closer look into specific datasets}

To illustrate changes in the performance of the compared algorithms, we further subsample the largest dataset in our suite -- mcf -- in order to obtain series of training sets of varying sizes from 0.01\% to 100\% of the original size.

We train \textsc{parrot-reuse} and \textsc{parrot-cache} on each of them, and then couple augmented algorithms.
The results are present in Figure~\ref{fig:results_mcf}.

As expected, the performance of the \textsc{parrot-cache} and \textsc{parrot-reuse} models improves along with the growth of the available data. We can observe that the performance of all the augmented algorithms continues to improve along with the performance of the coupled predictor. However, when the predictor underperforms \textsc{marker}, the performances of the augmented algorithms remain on par with the classical algorithm, with a notable exception of the Trust\&Doubt, which is clearly the worst.

However, as the \textsc{parrot-cache} and \textsc{parrot-reuse} predictors start to outperform \textsc{lru} and \textsc{marker}, the performance of the combining algorithms improves along with them. The most notable result here is how closely the RobustFtP$^\mathrm{D}$ follows improvements in the performance of \textsc{parrot-cache}, while overhead for remaining algorithms remains significantly larger.

At the first glance it might be surprising that sometimes, e.g. mcf (1\%), augmented algorithms outperform both \textsc{marker} and their coupled predictors.
However, an augmented algorithm can take advantage of the fact that each of its ingredients may perform better on a different part of the dataset.

Another dataset that we investigate in more detail is sphinx3. 
As it is smaller than mcf, we can only train models on the range from 0.1\% to 100\%.
Interestingly here on small subsamples (0.1\% and 1\%) the \textsc{parrot-reuse} performs better than \textsc{parrot-cache}, suggesting it might be more robust to scarcity of training data.
After careful examination the same trend can be observed comparing Figures~\ref{fig:results_full} and \ref{fig:results_1p}.

Since even the weakest predictors easily outperform \textsc{marker} on this dataset, the learning-augmented algorithms perform similar to what we see already in Section~\ref{sec:full}.

\subsection{Towards performance explanation}
\label{sec:usage}

\begin{figure}
    \centering
    \vspace{-0.3cm}
    \includegraphics[width=\columnwidth]{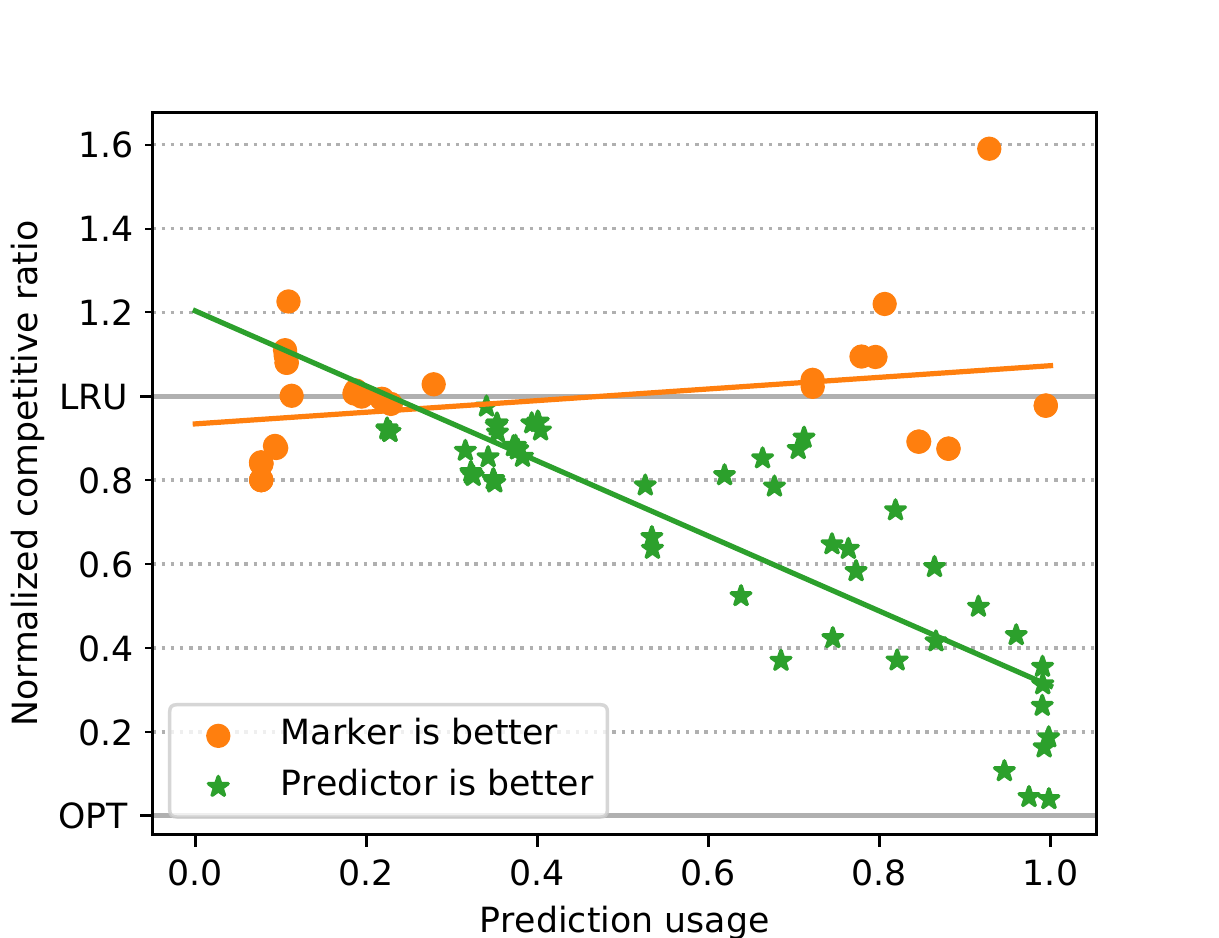}
    \vspace{-0.4cm}
    \caption{{\bf Prediction usage}. Each data point represents a pair (dataset, learning-augmented algorithm). Orange dots correspond to pairs such that \textsc{marker} performed better on that dataset than the predictor used by that algorithm (if followed blindly). Green stars correspond to pairs where the predictor was better. See Section~\ref{sec:usage} for the definition of \emph{prediction usage}. Intuitively, a good learning-augmented algorithm should have a large prediction usage for green stars and a small one for orange dots.}
    \label{fig:usage}

    \vspace{0.5cm}
    \includegraphics[width=\columnwidth]{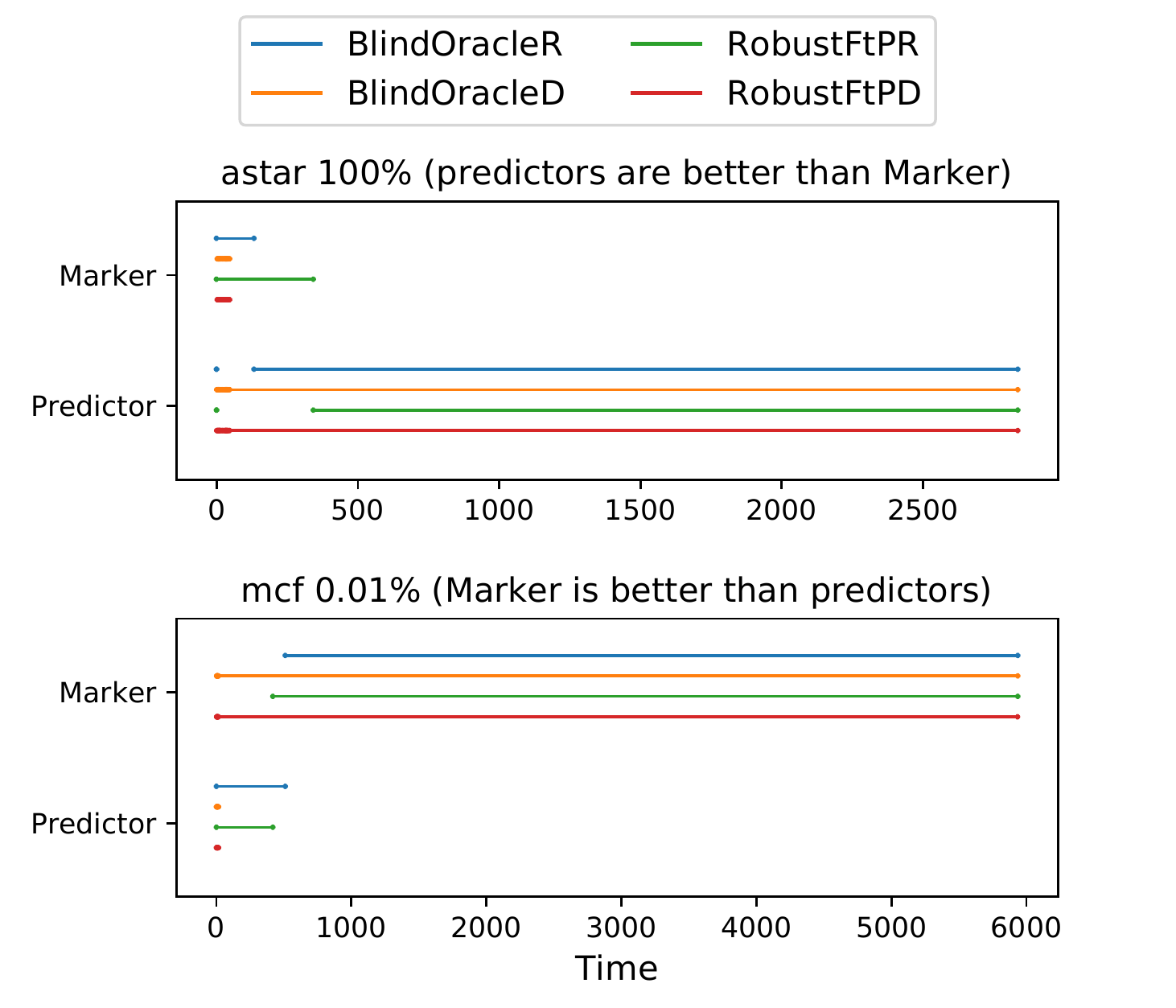}
    \vspace{-0.4cm}
    \caption{{\bf Switching behavior of the combiners}. For each time step, we draw a point in the upper part of the subplot if the combiner follows \textsc{marker}, and in the lower part if it follows the predictor. For astar 100\% dataset, the deterministic combiner, after a few switches, quickly infers that the predictor is better than \textsc{marker} and follows it until the end. On the other hand, the randomized combiner needs more time to make a decision. As a result, even though it follows the predictor most of the time, the initial hesitancy jeopardizes its overall performance. For mcf 0.01\% dataset, where the predictors are outperformed by \textsc{marker}, the deterministic combiner again outperforms randomized variant, as it quickly identifies what to follow.}
    \label{fig:switching}
\end{figure}
 
To better understand the differences in performance between the algorithms, we ran additional experiments (on a subset of datasets) in which we measured how much the algorithms followed the predictors. That notion is straightforward for BlindOracle and RobustFtP, as they, in each time step, either do exactly what the predictor advises, or follow \textsc{marker}, which in turn is completely independent of the predictor. However, the notion becomes more subtle for other algorithms, which, e.g., apply predictions only to a varying subset of items. To overcome that difficulty, we used a measure of \emph{prediction usage}, which is algorithm independent. Specifically, for each algorithm, we computed the Jaccard similarity between the caches maintained by the algorithm and the predictor (if followed blindly), averaged over time.

Our general conclusion is that algorithms' performance is correlated with how much they choose to follow the underlying predictor. The correlation is positive on datasets where the respective predictor performs better than \textsc{marker}, and negative otherwise, see Figure~\ref{fig:usage}. Most algorithms correctly decide for each dataset whether it is on average better to follow the predictor or not. It is, however, the level of commitment to that decision that differs: With fully-fledged predictors, deterministic BlindOracle$^\mathrm{D}$ and RobustFtP$^\mathrm{D}$ followed them \textgreater97\% of the time, randomized BlindOracle$^\mathrm{R}$ and RobustFtP$^\mathrm{R}$ -- around 90\% of the time, and remaining algorithms -- at most 85\%, often much less. For most algorithms, how much the algorithm follows the predictor seems to depend on the predictor's performance -- the better the predictor, the more it is followed. Only for Trust\&Doubt these two numbers seem uncorrelated, which may explain its lower robustness.

Next, we analyzed the switching behavior of the combiners, see Figure~\ref{fig:switching}. 
The deterministic combiner, after few brief switches, quickly infers what to follow for each dataset. The randomized combiner also eventually follows the better of the two underlying algorithms but needs much more time to figure out that it has to switch, which jeopardizes its overall performance. That explains why randomized BlindOracle$^{\mathrm{R}}$ and RobustFtP$^{\mathrm{R}}$ tend to perform worse than their deterministic counterparts, BlindOracle$^{\mathrm{D}}$ and RobustFtP$^{\mathrm{D}}$.

\section{Related Work}

\paragraph{Learning-augmented algorithms.}

The idea of augmenting an algorithm with hints or predictions coming from a potentially untrusted oracle is not new. The recent variant, clearly inspired by the now omnipresent machine-learned predictors for various tasks, seems to spark from \citet{LykourisV18} and \citet{PurohitSK18}. The idea has been applied to many problems, also beyond the online algorithms, e.g.~to Bloom filters~\cite{KraskaBCDP18}. For an overview of the field, see the recent survey by \citet{MV21}.

\paragraph{Robust machine learning.}

Robustness of machine learning methods is sometimes defined as robustness to \emph{adversarial examples} -- approximately estimated worst-case inputs laying controllably far from training distribution~\cite{carlini2017towards, weng2018proven,SzegedyZSBEGF13}.
More broadly, however, robustness in machine learning can be seen as an ability to generalize, i.e., to perform well on unseen examples~\cite{bishop2006pattern}, where a~distribution shift between training and testing examples poses a~challenge~\cite{Moreno-TorresRACH12,geirhos2020shortcut}.

We experiment with heterogeneous sequential datasets~\citep{henning2006,CRC17}.
Their characteristics change over time, as they are generated by real-world programs.
We leverage this property to test generalization ability -- and hence robustness -- of state-of-the-art machine learning predictors~\citep{LiuHSRA20}.
We vary amount of data available during training to analyze pessimistic cases and use learning-augmented algorithms to incorporate robustness to caching policies based on neural network predictions.

\section{Conclusions}

We fill a critical gap in the learning-augmented literature.
We evaluated the learning-augmented caching algorithms using the state-of-the-art predictors on real-world datasets.
In a nutshell, we conclude that learning-augmented algorithms can have only a~low overhead over a~well-performing predictor, while competing with classical methods when the coupled predictor fails, thus providing a~cheap worst-case insurance.

Our experiments show that when the training data is scarce, the performance of the state-of-the-art Parrot model tends to degrade quickly, depending on the dataset.
Hence, it justifies looking for a way to benefit from the robustness of the classical online algorithms. 
As learning augmented algorithms do exactly that, we test their performance in practice.

Two algorithms -- BlindOracle$^{\mathrm{D}}$ and RobustFtP$^{\mathrm{D}}$ -- turn out to be the best.
They provide a very low overhead over good predictions but still compete with the robust classical methods even when the predictors fail.
The remaining four tested algorithms are robust, but they seem to be overly conservative and do not fully utilize good predictions.

We show that the theoretical asymptotic competitive ratio is not a good proxy for the practical performance of the learning-augmented algorithms.
In the reuse distance setup, it correctly points to the leader but incorrectly distinguishes between the remaining algorithms. 
In the policy setup, the theoretically inferior algorithm turns out the best in practice.
Moreover, according to the theoretical analysis, the randomized combiner should perform better than the deterministic one, while in our experiments we observe the opposite.

On most datasets, the predictor for the optimal policy setup outperforms the predictor for the reuse distance setup. 
Hence, learning-augmented algorithms in the latter setup can hardly compete with the RobustFtP$^{\mathrm{D}}$, designed for the former.
That conclusion is valid with respect to current state-of-the-art predictors, and future improvements to reuse distance predictors may invalidate it.
However, in the view of our results, a direct empirical comparison between the two alternative setups becomes less relevant. Indeed, theoretical developments in both setups independently led to the same algorithmic idea behind BlindOracle$^{\mathrm{D}}$ and RobustFtP$^{\mathrm{D}}$. The idea excels in practice and can presumably be applied to any type of predictor. As the examples of BlindOracle$^{\mathrm{R}}$, RobustFtP$^{\mathrm{R}}$, and Trust\&Doubt show, optimizing the methods further towards an objective specific to a particular setup does not necessarily lead to improved performance \mbox{in~practice}.

\section*{Acknowledgements}
Adam Polak was supported by the Swiss National Science Foundation (SNF) within the project \emph{Lattice Algorithms and Integer Programming (185030)}.
Konrad Żołna was supported by the National Science Center, Poland (2017/27/N/ST6/00828, 2018/28/T/ST6/00211).

\bibliography{main}
\bibliographystyle{icml2021}

\clearpage

\appendix

\section{Supplementary material}

\paragraph{Additional information about experiments.}
We ran the total of 53 trainings and evaluations. All but two runs described below were run for $20\,000$ steps, with evaluation every $5000$ steps.

For datasets from \{astar, bwaves, bzip, cactusadm, gems, leslie3d, omnetpp, sphinx3\}, we ran 4 experiments on the first 0.1\%, 1\%, 10\% and 100\% records of the dataset.

For datasets from \{lbm, libq, milc, xalanc\} we ran only two experiments, on 1\% and 100\% of the dataset.

For the mcf dataset, we ran 11 experiments in total. 
The first five of them were runs on 0.01\%, 0.1\%, 1\%, 10\% and 100\% of the dataset, without DAgger.
For comparison, we ran additional six experiments with a larger number of steps and with DAgger. 
We achieved comparable cache hit rates, and therefore we did not evaluate learning-augmented algorithms on that additional data.

On average, each experiment took around one day on a Tesla V100 GPU, occupying at most 3GB of the GPU memory.

\paragraph{Full results.}
We present the full cache hit rates of \textsc{parrot-reuse} across all evaluated datasets and fractions in Table~\ref{hit-rates}. The raw unnormalized competitive ratios are shown in Table~\ref{comp-table}.

\begin{table*}[p]
\caption{\textbf{Cache hit rates of \textsc{parrot-cache}.} Detailed results.}
\label{hit-rates}
\begin{center}
\begin{small}
\begin{sc}
\resizebox{\textwidth}{!}{\begin{tabular}{lr|crrrrrrrrrr}
\textbf{Dataset} & \multicolumn{1}{r}{\textbf{Fraction}} & \rot{\textbf{Dagger}} & \multicolumn{1}{c}{\rot{\textbf{Steps}}} &
\multicolumn{1}{c}{\rot{\textbf{Eval freq}}} & \multicolumn{1}{c}{\rot{\textbf{Cache requests}}} & \multicolumn{1}{c}{\rot{\textbf{Cache requests in train set}}} & \multicolumn{1}{c}{\rot{\textbf{Hit Rate ckpt 0 on val}}} & \multicolumn{1}{c}{\rot{\textbf{Best non 0 checkpoint}}} & \multicolumn{1}{c}{\rot{\textbf{Hit Rate ckpt best on val}}} & \multicolumn{1}{c}{\rot{\textbf{Last checkpoint}}} & \multicolumn{1}{c}{\rot{\textbf{Hit Rate ckpt last on val}}} & \multicolumn{1}{c}{\rot{\textbf{Hit Rate ckpt best non 0 on test}}} \\
\midrule
astar & 0.1\% & FALSE & 20001 & 5000 & 640,032 & 1,154 & 8.25\% & 5000 & 13.32\% & 20000 & 11.83\% & 14.05\% \\
astar & 1\% & FALSE & 20001 & 5000 & 640,032 & 11,540 & 8.25\% & 5000 & 25.09\% & 20000 & 24.55\% & 25.75\% \\
astar & 10\% & FALSE & 20001 & 5000 & 640,032 & 115,404 & 8.25\% & 10000 & 30.26\% & 20000 & 30.05\% & 30.99\% \\
astar & 100\% & FALSE & 20001 & 5000 & 640,032 & 1,154,048 & 8.25\% & 20000 & 31.53\% & 20000 & 31.53\% & 32.17\% \\
\midrule
bwaves & 0.1\% & FALSE & 20001 & 5000 & 640,032 & 570 & 0.02\% & 5000 & 0.03\% & 20000 & 0.02\% & 0.00\% \\
bwaves & 1\% & FALSE & 20001 & 5000 & 640,032 & 5,703 & 0.02\% & 10000 & 1.93\% & 20000 & 1.62\% & 0.30\% \\
bwaves & 10\% & FALSE & 20001 & 5000 & 640,032 & 57,036 & 0.02\% & 20000 & 0.03\% & 20000 & 0.03\% & 0.00\% \\
bwaves & 100\% & FALSE & 20001 & 5000 & 640,032 & 570,368 & 0.02\% & 10000 & 3.21\% & 20000 & 1.21\% & 0.32\% \\
\midrule
bzip & 0.1\% & FALSE & 20001 & 5000 & 640,032 & 167 & 52.05\% & 20000 & 48.02\% & 20000 & 48.02\% & 53.23\% \\
bzip & 1\% & FALSE & 20001 & 5000 & 640,032 & 1,676 & 52.05\% & 5000 & 42.69\% & 20000 & 41.94\% & 45.12\% \\
bzip & 10\% & FALSE & 20001 & 5000 & 640,032 & 16,768 & 52.05\% & 5000 & 57.09\% & 20000 & 52.84\% & 64.28\% \\
bzip & 100\% & FALSE & 20001 & 5000 & 640,032 & 167,680 & 52.05\% & 5000 & 63.50\% & 20000 & 61.29\% & 68.44\% \\
\midrule
cactusadm & 0.1\% & FALSE & 20001 & 5000 & 640,032 & 221 & 0.32\% & 20000 & 4.81\% & 20000 & 4.81\% & 0.45\% \\
cactusadm & 1\% & FALSE & 20001 & 5000 & 640,032 & 2,219 & 0.32\% & 15000 & 8.43\% & 20000 & 8.42\% & 20.16\% \\
cactusadm & 10\% & FALSE & 20001 & 5000 & 640,032 & 22,195 & 0.32\% & 5000 & 25.88\% & 20000 & 24.17\% & 30.93\% \\
cactusadm & 100\% & FALSE & 20001 & 5000 & 640,032 & 221,952 & 0.32\% & 5000 & 25.30\% & 20000 & 24.98\% & 32.97\% \\
\midrule
gems & 0.1\% & FALSE & 20001 & 5000 & 640,032 & 723 & 2.96\% & 15000 & 3.14\% & 20000 & 3.11\% & 3.09\% \\
gems & 1\% & FALSE & 20001 & 5000 & 640,032 & 7,234 & 2.96\% & 20000 & 3.12\% & 20000 & 3.12\% & 3.09\% \\
gems & 10\% & FALSE & 20001 & 5000 & 640,032 & 72,345 & 2.96\% & 20000 & 3.13\% & 20000 & 3.13\% & 3.08\% \\
gems & 100\% & FALSE & 20001 & 5000 & 640,032 & 723,456 & 2.96\% & 20000 & 1.77\% & 20000 & 1.77\% & 3.10\% \\
\midrule
lbm & 1\% & FALSE & 20001 & 5000 & 640,032 & 7,820 & 0.02\% & 5000 & 0.19\% & 20000 & 0.13\% & 0.00\% \\
lbm & 100\% & FALSE & 20001 & 5000 & 640,032 & 782,080 & 0.02\% & 5000 & 1.22\% & 20000 & 0.29\% & 0.00\% \\
\midrule
leslie3d & 0.1\% & FALSE & 20001 & 5000 & 640,032 & 716 & 0.81\% & 15000 & 1.29\% & 20000 & 1.01\% & 9.10\% \\
leslie3d & 1\% & FALSE & 20001 & 5000 & 640,032 & 7,160 & 0.81\% & 5000 & 1.62\% & 20000 & 1.21\% & 9.31\% \\
leslie3d & 10\% & FALSE & 20001 & 5000 & 640,032 & 71,603 & 0.81\% & 5000 & 2.09\% & 20000 & 1.00\% & 11.73\% \\
leslie3d & 100\% & FALSE & 20001 & 5000 & 640,032 & 716,032 & 0.81\% & 5000 & 4.99\% & 20000 & 4.89\% & 11.41\% \\
\midrule
libq & 1\% & FALSE & 20001 & 5000 & 640,032 & 5,798 & 0.00\% & 5000 & 0.00\% & 20000 & 0.00\% & 0.00\% \\
libq & 100\% & FALSE & 20001 & 5000 & 640,032 & 579,840 & 0.00\% & 10000 & 0.01\% & 20000 & 0.00\% & 0.01\% \\
\midrule
mcf & 0.01\% & FALSE & 20001 & 5000 & 640,032 & 296 & 2.61\% & 5000 & 15.79\% & 20000 & 11.02\% & 15.27\% \\
mcf & 0.01\% & TRUE & 20001 & 5000 & 640,032 & 296 & 2.61\% & 5000 & 15.79\% & 20000 & 11.02\% & 15.28\% \\
mcf & 0.1\% & FALSE & 20001 & 5000 & 640,032 & 2,965 & 2.61\% & 15000 & 22.19\% & 20000 & 22.16\% & 27.33\% \\
mcf & 0.1\% & TRUE & 20001 & 5000 & 640,032 & 2,965 & 2.61\% & 10000 & 21.64\% & 20000 & 21.40\% & 27.02\% \\
mcf & 1\% & FALSE & 20001 & 5000 & 640,032 & 29,655 & 2.61\% & 5000 & 40.00\% & 20000 & 39.44\% & 41.31\% \\
mcf & 1\% & TRUE & 20001 & 5000 & 640,032 & 29,655 & 2.61\% & 5000 & 40.35\% & 20000 & 39.93\% & 41.42\% \\
mcf & 10\% & FALSE & 20001 & 5000 & 640,032 & 296,550 & 2.61\% & 5000 & 42.09\% & 20000 & 41.49\% & 43.45\% \\
mcf & 100\% & FALSE & 20001 & 5000 & 640,032 & 2,965,504 & 2.61\% & 20000 & 43.05\% & 20000 & 43.05\% & 43.89\% \\
mcf & 100\% & TRUE & 20001 & 5000 & 640,032 & 2,965,504 & 2.61\% & 15000 & 42.65\% & 20000 & 42.49\% & 43.46\% \\
mcf & 100\% & FALSE & 120001 & 30000 & 3,840,032 & 2,965,504 & 2.61\% & 30000 & 43.10\% & 120000 & 43.10\% & 43.89\% \\
mcf & 100\% & TRUE & 120001 & 30000 & 3,840,032 & 2,965,504 & 2.61\% & 30000 & 42.60\% & 120000 & 42.60\% & 43.11\% \\
\midrule
milc & 1\% & FALSE & 20001 & 5000 & 640,032 & 5,568 & 0.22\% & 10000 & 0.23\% & 20000 & 0.23\% & 0.01\% \\
milc & 100\% & FALSE & 20001 & 5000 & 640,032 & 556,800 & 0.22\% & 5000 & 0.34\% & 20000 & 0.23\% & 0.01\% \\
\midrule
omnetpp & 0.1\% & FALSE & 20001 & 5000 & 640,032 & 555 & 8.24\% & 5000 & 16.92\% & 20000 & 16.60\% & 17.90\% \\
omnetpp & 1\% & FALSE & 20001 & 5000 & 640,032 & 5,555 & 8.24\% & 10000 & 13.85\% & 20000 & 13.62\% & 16.85\% \\
omnetpp & 10\% & FALSE & 20001 & 5000 & 640,032 & 55,552 & 8.24\% & 5000 & 18.60\% & 20000 & 14.83\% & 19.28\% \\
omnetpp & 100\% & FALSE & 20001 & 5000 & 640,032 & 555,520 & 8.24\% & 5000 & 22.23\% & 20000 & 15.46\% & 21.91\% \\
\midrule
sphinx3 & 0.1\% & FALSE & 20001 & 5000 & 640,032 & 328 & 36.60\% & 20000 & 36.33\% & 20000 & 36.33\% & 53.78\% \\
sphinx3 & 1\% & FALSE & 20001 & 5000 & 640,032 & 3,287 & 36.60\% & 5000 & 43.84\% & 20000 & 42.71\% & 59.48\% \\
sphinx3 & 10\% & FALSE & 20001 & 5000 & 640,032 & 32,870 & 36.60\% & 5000 & 45.68\% & 20000 & 45.49\% & 66.55\% \\
sphinx3 & 100\% & FALSE & 20001 & 5000 & 640,032 & 328,704 & 36.60\% & 20000 & 55.18\% & 20000 & 55.18\% & 70.54\% \\
\midrule
xalanc & 1\% & FALSE & 20001 & 5000 & 640,032 & 691 & 14.73\% & 15000 & 15.02\% & 20000 & 14.46\% & 32.38\% \\
xalanc & 100\% & FALSE & 20001 & 5000 & 640,032 & 69,120 & 14.73\% & 5000 & 27.01\% & 20000 & 26.42\% & 49.71\%
\end{tabular}}
\end{sc}
\end{small}
\end{center}

\end{table*}

\begin{table*}[p]
\caption{\textbf{Competitive ratios.} Detailed results.}
\label{comp-table}
\begin{center}
\begin{small}
\begin{sc}
\resizebox{\textwidth}{!}{\begin{tabular}{lr|ccccccccccccccc}
\textbf{Dataset} & \multicolumn{1}{r}{\textbf{Fraction}} & \rot{\textbf{OPT}} & \rot{\textbf{LRU}} & \rot{\textbf{Marker}} & \rot{\textbf{Random}} & \rot{\textbf{Parrot-Reuse}} & \rot{\textbf{PredictiveMarker}} & \rot{\textbf{Lmarker}} & \rot{\textbf{LNonMarkerR}} & \rot{\textbf{LNonMarkerD}} & \rot{\textbf{BlindOracleR}} & \rot{\textbf{BlindOracleD}} & \rot{\textbf{Parrot-Cache}} & \rot{\textbf{Trust\&Doubt}} & \rot{\textbf{RobustFtPR}} & \rot{\textbf{RobustFtPD}} \\
\midrule
astar & 0.1\% & 1.00 & 1.53 & 1.52 & 1.47 & 1.26 & 1.51 & 1.51 & 1.51 & 1.51 & 1.32 & 1.27 & 1.37 & 1.53 & 1.42 & 1.38 \\
astar & 1\% & 1.00 & 1.53 & 1.52 & 1.47 & 1.16 & 1.50 & 1.50 & 1.50 & 1.49 & 1.22 & 1.17 & 1.19 & 1.47 & 1.24 & 1.19 \\
astar & 10\% & 1.00 & 1.53 & 1.52 & 1.47 & 1.15 & 1.50 & 1.50 & 1.50 & 1.49 & 1.19 & 1.15 & 1.10 & 1.45 & 1.14 & 1.11 \\
astar & 100\% & 1.00 & 1.53 & 1.52 & 1.46 & 1.14 & 1.50 & 1.50 & 1.50 & 1.49 & 1.19 & 1.14 & 1.08 & 1.45 & 1.14 & 1.09 \\
\midrule
bwaves & 0.1\% & 1.00 & 1.05 & 1.05 & 1.05 & 1.05 & 1.05 & 1.05 & 1.05 & 1.05 & 1.05 & 1.05 & 1.05 & 1.05 & 1.05 & 1.05 \\
bwaves & 1\% & 1.00 & 1.05 & 1.05 & 1.05 & 1.05 & 1.05 & 1.05 & 1.05 & 1.05 & 1.05 & 1.05 & 1.05 & 1.05 & 1.05 & 1.05 \\
bwaves & 10\% & 1.00 & 1.05 & 1.05 & 1.05 & 1.05 & 1.05 & 1.05 & 1.05 & 1.05 & 1.05 & 1.05 & 1.05 & 1.05 & 1.05 & 1.05 \\
bwaves & 100\% & 1.00 & 1.05 & 1.05 & 1.05 & 1.05 & 1.05 & 1.05 & 1.05 & 1.05 & 1.05 & 1.05 & 1.05 & 1.05 & 1.05 & 1.05 \\
\midrule
bzip & 0.1\% & 1.00 & 1.88 & 1.93 & 2.23 & 2.96 & 2.00 & 1.98 & 2.00 & 1.97 & 2.36 & 1.92 & 2.43 & 2.00 & 2.17 & 1.98 \\
bzip & 1\% & 1.00 & 1.88 & 1.96 & 2.22 & 2.92 & 1.98 & 1.97 & 1.98 & 1.98 & 2.26 & 1.92 & 2.85 & 2.03 & 2.32 & 2.00 \\
bzip & 10\% & 1.00 & 1.88 & 1.93 & 2.26 & 2.39 & 1.93 & 1.93 & 1.97 & 1.95 & 2.12 & 1.92 & 1.86 & 1.81 & 1.90 & 1.84 \\
bzip & 100\% & 1.00 & 1.88 & 1.94 & 2.26 & 2.19 & 1.94 & 1.94 & 1.98 & 1.96 & 2.06 & 1.92 & 1.64 & 1.69 & 1.78 & 1.66 \\
\midrule
cactusadm & 0.1\% & 1.00 & 1.51 & 1.49 & 1.44 & 1.46 & 1.47 & 1.47 & 1.48 & 1.47 & 1.48 & 1.47 & 1.50 & 1.51 & 1.49 & 1.50 \\
cactusadm & 1\% & 1.00 & 1.51 & 1.49 & 1.44 & 1.36 & 1.46 & 1.46 & 1.47 & 1.47 & 1.42 & 1.37 & 1.20 & 1.46 & 1.34 & 1.22 \\
cactusadm & 10\% & 1.00 & 1.51 & 1.49 & 1.44 & 1.19 & 1.45 & 1.45 & 1.47 & 1.45 & 1.33 & 1.20 & 1.04 & 1.40 & 1.21 & 1.05 \\
cactusadm & 100\% & 1.00 & 1.51 & 1.49 & 1.44 & 1.18 & 1.44 & 1.44 & 1.46 & 1.45 & 1.31 & 1.19 & 1.01 & 1.40 & 1.15 & 1.02 \\
\midrule
gems & 0.1\% & 1.00 & 1.11 & 1.09 & 1.09 & 1.13 & 1.09 & 1.09 & 1.09 & 1.09 & 1.11 & 1.11 & 1.10 & 1.10 & 1.10 & 1.09 \\
gems & 1\% & 1.00 & 1.11 & 1.09 & 1.10 & 1.13 & 1.09 & 1.09 & 1.09 & 1.09 & 1.11 & 1.11 & 1.10 & 1.10 & 1.10 & 1.09 \\
gems & 10\% & 1.00 & 1.11 & 1.09 & 1.09 & 1.13 & 1.08 & 1.09 & 1.09 & 1.09 & 1.11 & 1.11 & 1.10 & 1.10 & 1.10 & 1.09 \\
gems & 100\% & 1.00 & 1.11 & 1.09 & 1.10 & 1.13 & 1.09 & 1.09 & 1.09 & 1.09 & 1.11 & 1.11 & 1.10 & 1.10 & 1.10 & 1.09 \\
\midrule
lbm & 1\% & 1.00 & 1.33 & 1.33 & 1.30 & 1.32 & 1.33 & 1.33 & 1.33 & 1.33 & 1.33 & 1.33 & 1.33 & 1.33 & 1.33 & 1.33 \\
lbm & 100\% & 1.00 & 1.33 & 1.33 & 1.30 & 1.32 & 1.33 & 1.33 & 1.33 & 1.33 & 1.33 & 1.33 & 1.33 & 1.33 & 1.33 & 1.33 \\
\midrule
leslie3d & 0.1\% & 1.00 & 1.31 & 1.31 & 1.31 & 1.39 & 1.32 & 1.32 & 1.32 & 1.32 & 1.34 & 1.32 & 1.32 & 1.32 & 1.31 & 1.31 \\
leslie3d & 1\% & 1.00 & 1.31 & 1.31 & 1.31 & 1.39 & 1.31 & 1.31 & 1.31 & 1.31 & 1.34 & 1.32 & 1.31 & 1.31 & 1.31 & 1.31 \\
leslie3d & 10\% & 1.00 & 1.31 & 1.31 & 1.31 & 1.37 & 1.31 & 1.31 & 1.31 & 1.31 & 1.33 & 1.32 & 1.28 & 1.30 & 1.29 & 1.28 \\
leslie3d & 100\% & 1.00 & 1.31 & 1.31 & 1.31 & 1.38 & 1.31 & 1.31 & 1.31 & 1.31 & 1.33 & 1.32 & 1.28 & 1.30 & 1.30 & 1.29 \\
\midrule
libq & 1\% & 1.00 & 1.06 & 1.06 & 1.06 & 1.00 & 1.06 & 1.06 & 1.06 & 1.06 & 1.02 & 1.01 & 1.06 & 1.06 & 1.06 & 1.06 \\
libq & 100\% & 1.00 & 1.06 & 1.06 & 1.06 & 1.00 & 1.06 & 1.06 & 1.06 & 1.06 & 1.03 & 1.01 & 1.06 & 1.06 & 1.06 & 1.06 \\
\midrule
mcf & 0.01\% & 1.00 & 1.32 & 1.36 & 1.43 & 1.72 & 1.34 & 1.34 & 1.35 & 1.35 & 1.37 & 1.32 & 1.53 & 1.50 & 1.36 & 1.34 \\
mcf & 0.1\% & 1.00 & 1.32 & 1.35 & 1.43 & 1.58 & 1.31 & 1.32 & 1.33 & 1.32 & 1.37 & 1.32 & 1.31 & 1.26 & 1.29 & 1.28 \\
mcf & 1\% & 1.00 & 1.32 & 1.35 & 1.43 & 1.29 & 1.26 & 1.26 & 1.28 & 1.26 & 1.27 & 1.26 & 1.06 & 1.21 & 1.08 & 1.06 \\
mcf & 10\% & 1.00 & 1.32 & 1.36 & 1.43 & 1.22 & 1.25 & 1.25 & 1.27 & 1.24 & 1.21 & 1.20 & 1.02 & 1.20 & 1.05 & 1.02 \\
mcf & 100\% & 1.00 & 1.32 & 1.36 & 1.43 & 1.22 & 1.25 & 1.25 & 1.26 & 1.25 & 1.22 & 1.20 & 1.01 & 1.20 & 1.03 & 1.01 \\
\midrule
milc & 1\% & 1.00 & 1.01 & 1.01 & 1.01 & 1.00 & 1.01 & 1.01 & 1.01 & 1.01 & 1.01 & 1.01 & 1.01 & 1.01 & 1.01 & 1.01 \\
milc & 100\% & 1.00 & 1.01 & 1.01 & 1.01 & 1.00 & 1.01 & 1.01 & 1.01 & 1.01 & 1.01 & 1.01 & 1.01 & 1.01 & 1.01 & 1.01 \\
\midrule
omnetpp & 0.1\% & 1.00 & 1.38 & 1.39 & 1.43 & 1.56 & 1.38 & 1.38 & 1.39 & 1.39 & 1.44 & 1.39 & 1.43 & 1.39 & 1.40 & 1.39 \\
omnetpp & 1\% & 1.00 & 1.38 & 1.39 & 1.43 & 1.56 & 1.38 & 1.38 & 1.39 & 1.39 & 1.44 & 1.39 & 1.44 & 1.42 & 1.41 & 1.40 \\
omnetpp & 10\% & 1.00 & 1.38 & 1.39 & 1.43 & 1.55 & 1.38 & 1.38 & 1.39 & 1.38 & 1.45 & 1.39 & 1.40 & 1.40 & 1.40 & 1.39 \\
omnetpp & 100\% & 1.00 & 1.38 & 1.39 & 1.43 & 1.53 & 1.38 & 1.38 & 1.39 & 1.38 & 1.46 & 1.39 & 1.36 & 1.37 & 1.37 & 1.35 \\
\midrule
sphinx3 & 0.1\% & 1.00 & 3.46 & 2.29 & 1.87 & 1.43 & 1.56 & 1.59 & 1.93 & 1.68 & 1.69 & 1.44 & 1.83 & 2.09 & 2.00 & 1.86 \\
sphinx3 & 1\% & 1.00 & 3.46 & 2.30 & 1.88 & 1.40 & 1.84 & 1.92 & 1.98 & 1.85 & 1.62 & 1.41 & 1.60 & 1.87 & 1.82 & 1.62 \\
sphinx3 & 10\% & 1.00 & 3.46 & 2.30 & 1.87 & 1.36 & 1.58 & 1.65 & 1.89 & 1.66 & 1.59 & 1.37 & 1.32 & 1.63 & 1.69 & 1.33 \\
sphinx3 & 100\% & 1.00 & 3.46 & 2.3 & 1.88 & 1.28 & 1.52 & 1.56 & 1.87 & 1.61 & 1.63 & 1.29 & 1.16 & 1.47 & 1.46 & 1.17 \\
\midrule
xalanc & 1\% & 1.00 & 1.27 & 1.32 & 1.48 & 1.87 & 1.34 & 1.34 & 1.34 & 1.33 & 1.61 & 1.30 & 1.56 & 1.37 & 1.42 & 1.34 \\
xalanc & 100\% & 1.00 & 1.27 & 1.32 & 1.47 & 1.45 & 1.29 & 1.28 & 1.31 & 1.30 & 1.38 & 1.29 & 1.17 & 1.17 & 1.23 & 1.19
\end{tabular}}
\end{sc}
\end{small}
\end{center}
\end{table*}

\end{document}